\documentclass[final,5p,times,twocolumn]{elsarticle}

\usepackage{amsmath,amssymb,amsfonts}
\usepackage{bm}
\usepackage{multirow}
\usepackage{makecell}
\usepackage{arydshln}
\usepackage{url}

\begin{document}

\begin{frontmatter}



\title{Equipping Sketch Patches with Context-Aware Positional Encoding for Graphic Sketch Representation}


\author[label1]{Sicong Zang\corref{cor1}}
\ead{sczang@dhu.edu.cn}
\cortext[cor1]{Corresponding author}

\author[label1]{Zhijun Fang}
\ead{zjfang@dhu.edu.cn}

\affiliation[label1]{organization={School of Computer Science and Technology, Donghua University}, 
			city={Shanghai}, 
			postcode={201620},
			country={China}}

\begin{abstract}
When benefiting graphic sketch representation with sketch drawing orders, recent studies have linked sketch patches as graph edges by drawing orders in accordance to a temporal-based nearest neighboring strategy. However, such constructed graph edges may be unreliable, since the contextual relationships between patches may be inconsistent with the sequential positions in drawing orders, due to variants of sketch drawings. In this paper, we propose a variant-drawing-protected method by equipping sketch patches with context-aware positional encoding (PE) to make better use of drawing orders for sketch learning. We introduce a sinusoidal absolute PE to embed the sequential positions in drawing orders, and a learnable relative PE to encode the unseen contextual relationships between patches. Both types of PEs never attend the construction of graph edges, but are injected into graph nodes to cooperate with the visual patterns captured from patches. After linking nodes by semantic proximity, during message aggregation via graph convolutional networks, each node receives both semantic features from patches and contextual information from PEs from its neighbors, which equips local patch patterns with global contextual information, further obtaining drawing-order-enhanced sketch representations. Experimental results indicate that our method significantly improves sketch healing and controllable sketch synthesis. The source codes could be found at \url{https://github.com/SCZang/DC-gra2seq}.
\end{abstract}



\end{frontmatter}


\section{Introduction}\label{sec:intr}

Free-hand sketches carry vivid emotions and messages for communication through human history. The drawing order of a sketch records how it is drawn stroke-by-stroke by a human being. It provides models a unique view to review how sketch components are positioned in sequential orders, which fails to be stored on the canvas by sketch images seen in daily life. 

\begin{figure*}[ht]
	\centering
	\includegraphics[width=0.8\linewidth]{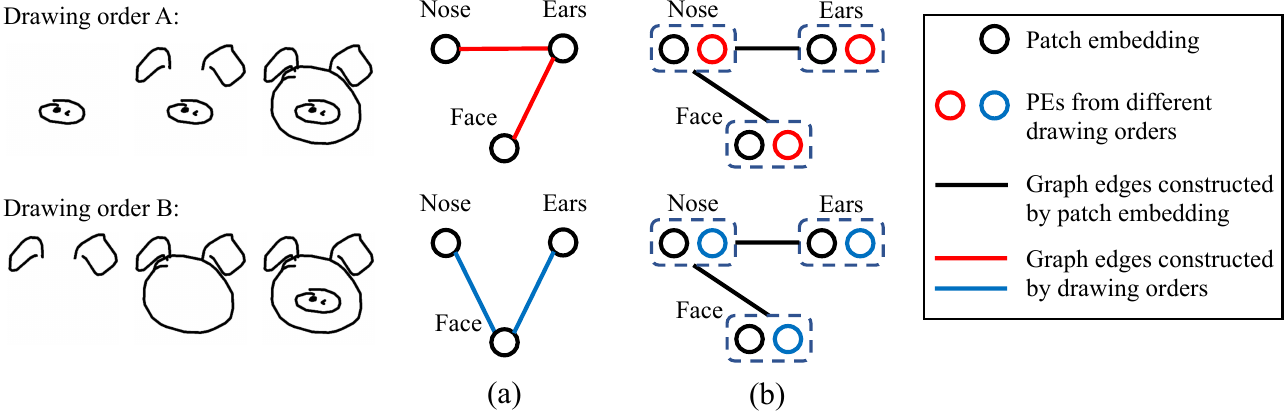}
	\caption{Two approaches to inject drawing orders into graphic sketch representation, when dealing with variants of sketch drawings. (a) Constructing graph edges by drawing orders as in \cite{su2020sketchhealer,qi2022generative}, by linking the neighboring sketch components following a drawing order. (b) The proposed approach by injecting drawing orders into graph edges, keeping drawing orders away from graph edge construction.}
	\label{fig:pe_for_node}
\end{figure*}

Recent studies have utilized sequential information from sketch drawing orders to cooperate with the visual patterns from sketch images for sketch learning. One common approach is employing a two-branch encoder to learn sketch representations from sketch images and their corresponding paired sketch sequence simultaneously \cite{song2018learning,xu2018sketchmate,xu2020learning,xu2020deep}. The combination of temporal and visual features is flexible and beneficial to learn accurate sketch representation.

Another effective way is injecting drawing orders into graphic sketch representation, e.g., constructing graph edges by drawing-order-based sequential positions \cite{su2020sketchhealer,qi2022generative}. However, as pointed out by \cite{zang2023linking}, due to variations of sketch drawings, the uncertainty of graph edge construction brought by different drawing orders, shown in Fig. \ref{fig:pe_for_node}(a), may interrupt message aggregation via graph convolutional network (GCN) layers \cite{kipf2016semi}, resulting in inaccurate sketch learning. Thus, it is doubtful whether the sequential information stored in drawing orders could reveal reliable contextual relationships between sketch components. Such sequential relationships in sketches are different from the contextual relationships between tokens in natural languages processing. Exchanging the positions of two tokenized phrases would corrupt a sentence to lose its semantic meaning. But for a sketch sequence, its drawing order never determines what it looks like on the canvas. As a result, the sequential positions in a drawing order might fail to precisely reveal their actual contextual relationships.

It motivates us to make better use of sketch drawing orders by injecting them into graph nodes instead of edges, shown in Fig. \ref{fig:pe_for_node}(b). More specifically, each sketch patch regarded as graph nodes are enriched by its sequential position in drawing orders encoded by a vectorized absolute positional encoding (PE) \cite{gehring2017convolutional,vaswani2017attention}. Absolute PE records when a patch is drawn in a drawing order\footnote{The cropping centers of sketch patches are selected following the drawing orders \cite{su2020sketchhealer,qi2022generative,zang2023linking}. Thus, we transplant the sequential positions in drawing order to rank patches.}. Moreover, in order to capture some hints about the unseen contextual relationships, we embed the ``contextual distances'' between semantically related patches along drawing orders, by introducing a learnable relative PE. With the equipment of relative PE, each patch enables to realize how far away its semantically related patches are in a drawing order. By the attendance of PEs, during message aggregation via GCN layers, the position-unaware graph nodes receives not only the visual contents from patches but also their contextual information, guiding our method to capture more hidden patterns from the unique view of drawing. Furthermore, our graph edges are constructed by semantic similarity between patches as in \cite{zang2023linking}, leaving drawing orders away from node linking to protect graph construction from variants of sketch drawings.

To realize the above proposals, we propose Drawing-order-enhanced Context-aware graph to sequence (DC-gra2seq) to enhance graphic sketch representation learning with some contextual information from drawing orders. Each patch embedding, captured by a convolutional neural network (CNN) encoder, is equipped with a sinusoidal absolute PE to highlight the sequential position in drawing orders. Moreover, its neighboring patches, which are ranked by the values of cosine similarity between patch embeddings, are offered with learnable relative PEs to restore the contextual positions within a neighborhood. Both patch embeddings and PEs are incorporated by a GCN encoder to obtain the final sketch representation, which is sent into a recurrent neural network (RNN) decoder for sketch generation. To summarize, we make the following contributions:
\begin{itemize}
	\item[1.] We propose DC-gra2seq to learn graphic sketch representation by introducing context-aware PEs to make better use of sketch drawing orders.
	\item[2.] We rethink the attendance of drawing orders in graphic sketch representation, and equip them on graph nodes to access some contextual information from drawing orders, but keep them away from graph edges to protect sketch learning from variants of sketch drawings.
	\item[3.] Experimental results indicate that DC-gra2seq achieves significant improvements on sketch healing and controllable sketch synthesis.
\end{itemize}

\section{Related work}

\subsection{Learning sketch representations}

Sketches are usually formed by raster images to restore visual patterns from the canvas, or formed by sequences of coordinates to record how they are drawn stroke-by-stroke by human. For sketch images, CNN-based methods, e.g., \cite{chen2017sketch,zang2021controllable,zang2024self,li2024lmser,agarwal2023sketchbuddy}, could be utilized for representation learning to capture the spatial dependencies among drawing strokes. For sketch sequences, in order to extract temporal relationships between strokes, RNN-based, e.g., \cite{ha2017neural,li2018universal}, or transformer-based methods, e.g., \cite{ribeiro2020sketchformer,lin2020sketch} have been proposed to learn the contextual information along sketch drawing orders. Moreover, using dual-branch encoders, e.g., \cite{song2018learning,xu2018sketchmate,xu2020deep,xu2020learning}, is beneficial to learn accurate and efficient latent representations from paired sketches (a sketch image and its corresponding sketch sequence), by making better use of both temporal and spatial relationships among sketch components.

Recently, graphic sketch representation is demonstrated as an effective way for sketch representation learning in sketch synthesis \cite{su2020sketchhealer,qi2022generative,qi2021sketchlattice,zang2023linking}, sketch recognition \cite{li2021efficient,xu2021multigraph,yang2020s}, sketch segmentation \cite{qi2022one,yang2021sketchgnn} and sketch-based image retrieval \cite{zhang2020zero}. A sketch is formed as a graph with multiple components regarded as graph nodes, whose captured features are aggregated by graph neural network (GNN) \cite{scarselli2008graph} or GCN layers to obtain the final sketch representation. Graph nodes can be patches cropped from sketch images \cite{su2020sketchhealer,qi2022generative,zang2023linking}, coordinates on a latticed canvas \cite{qi2021sketchlattice}, etc, carrying local sketch patterns. Graph edges can be constructed by linking sketch components to follow sketch drawing order \cite{su2020sketchhealer,qi2022generative}, Euclidean distances between sketch components \cite{qi2021sketchlattice,yang2021sketchgnn}, or semantic similarity \cite{zang2023linking}, etc. Graphic representation enables the message passing between sketch components for learning better sketch representation.

It is beneficial to highlight some valuable information in graph edges or nodes to assist graphic representation learning, e.g., constructing graph edges by the spatial relationships between image regions in a criss-cross way \cite{wang2020graph}. We aim to equip graph nodes with contextual relationships from sketch drawing order via PEs, which makes message aggregating process via GCN layers context-aware for learning accurate sketch representations.

\subsection{Positional encoding}

Positional encoding makes better use of some valuable sequential or structural information, when training position-unaware networks, e.g., self-attention network \cite{vaswani2017attention} and graph attention network \cite{velivckovic2017graph}. PEs give these architectures a sense of which portion of the sequence in input (or output) it is currently dealing with. Recent studies embedded the positional information by absolute \cite{gehring2017convolutional,vaswani2017attention}, relative PEs \cite{shaw2018self,dai2019transformer,wu2021rethinking} or both \cite{wang2019self}. Absolute PEs restore the absolute positions of the input tokens, indicating where a specific token is located in a sequence or a canvas. Relative PEs encode the relative pairwise distances or positions of a pair of tokens, which is invariant to the position of focal token in contrast to absolute PE. The positional information can be embedded by either fixed \cite{vaswani2017attention,wang2019self} or learnable PEs \cite{gehring2017convolutional,shaw2018self,wu2021rethinking}.

We inject sketch drawing orders into graph nodes by equipping sketch patches with both absolute and relative PEs. Each patch collects the visual contents from its neighboring patches along with their contextual relationships in drawing orders, during a message aggregating process.

\begin{figure*}[ht]
	\centering
	\includegraphics[width=\linewidth]{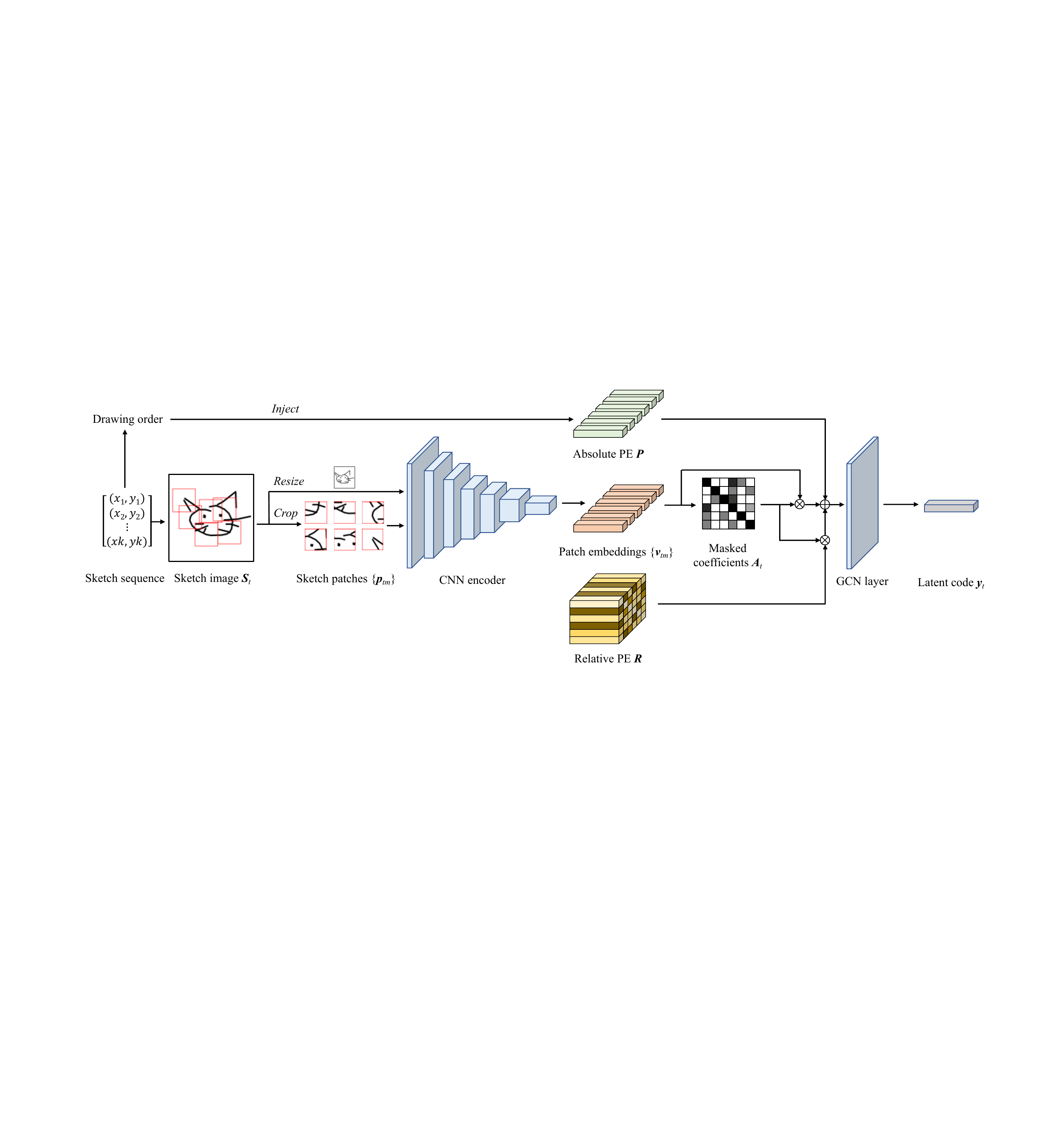}
	\caption{Learning graphic representation $\bm y_t$ of sketch $\bm S_t$ by DC-gra2seq. The cropped sketch patches $\{\bm p_{tm}\}$ along with the resized full sketch $\bm p_{t0}$ are embedded by a CNN encoder as patch embeddings $\{\bm v_{tm}\}$. The absolute PE $\bm P$ restoring sketch drawing order and the relative PE $\bm R$ encoding contextual relationships between patches are incorporated with $\{\bm v_{tm}\}$, weighted by masked coefficients $\bm A_t$ computed from $\{\bm v_{tm}\}$ only. A GCN layer collects all the information from $\bm v$, $\bm P$ and $\bm R$ to produce a final sketch code $\bm y_t$.}
	\label{fig:overview}
\end{figure*}

\section{Methodology}

Fig. \ref{fig:overview} offers an overview to indicate how DC-gra2seq learns graphic sketch representation. The embeddings of cropped sketch patches as graph nodes are captured by a CNN encoder, and the graph edges between them are constructed by masked coefficients computed from patch embeddings. During message aggregation via a GCN layer, both absolute and relative PEs are equipped on graph node to make use of the sequential and contextual relationships in drawing orders, producing the final sketch code, which is fed into an RNN decoder to reconstruct the input sketch in a sequence form.

\subsection{Encoding contextual information from sketch drawing order by PEs}

For the $t$-th sketch image $\bm S_t\in\mathbb{R}^{640\times 640}$ as input, we crop $M$ sketch patches $\{\bm p_{tm}\}_{m=1}^M$ with the size of $256\times 256$ from the canvas. We adopt the selecting rule utilized in \cite{su2020sketchhealer,qi2022generative,zang2023linking} to determine the cropping centers of patches, which arranges these patches in a sequential order in accordance to the drawing orders. More specifically, as the cropping center of $\bm p_{tm}$ is located on the stroke drawn before the one of $\bm p_{t,m+1}$, $\bm p_{tm}$ is positioned prior to $\bm p_{t,m+1}$ in the sequential order. Such sequential positions make patches position-aware, and we inject them into sinusoidal \textbf{absolute PE} \cite{vaswani2017attention}, shown in Eq. (\ref{eq:abs_pe}).
\begin{align}
	\label{eq:abs_pe}
	\bm P(pos, 2d)&=\sin\left(pos/10000^{2d/512}\right),\notag\\
	\bm P(pos, 2d+1)&=\cos\left(pos/10000^{2d/512}\right),
\end{align}
where $pos$ and $d$ denote the position and dimension, respectively. $512$ indicates the length of vectorized PE. $\bm P(m,\cdot)$ (abbreviated as $\bm P(m)$ in the following) indicates the absolute PE of $\bm p_{tm}$ to record when $\bm p_{tm}$ is drawn in a drawing order. Absolute PEs enrich the position-unaware sketch patches with sequential hints, which cannot be captured from canvas, and their attendances in sketch learning guide DC-gra2seq to consider both visual patterns and sequential information jointly.

Furthermore, as mentioned in Sect. \ref{sec:intr}, the contextual relationships between drawing strokes may fail to be revealed by the sequential positions in drawing orders, due to variants of sketch drawings. Thus, we introduce the \textbf{learnable relative PE} $\bm R(i, j)$ to capture the contextual distances between sketch patches along drawing orders. $\bm R(i, j)$ is a learnable vector with the same length of $\bm P(i)$, encoding the contextual relationships between patches $\bm p_{ti}$ and $\bm p_{tj}$. We restrict $\bm R(i, j)$ with the following two conditions:
\begin{align}
	\bm R(i, i+k)&=\bm R(j, j+k),\label{eq:target_invariant}\\
	\bm R(i, j)&=\bm R(j, i).\label{eq:undirected}
\end{align}

Firstly, our relative PE is \emph{target-invariant} by Eq. (\ref{eq:target_invariant}), i.e., relative PE is insensitive to the content of patches. Two pairs of patches with the same contextual distances, e.g., ($\bm p_{ti}$, $\bm p_{t, i+k}$) and ($\bm p_{tj}$ and $\bm p_{t, j+k}$), share the same value of relative PE. Secondly, our relative PE is \emph{undirected} by Eq. (\ref{eq:undirected}), i.e., the sequential order of two patches are not counted. $\bm R(i, j)$ only encodes how far away two patches $\bm p_{ti}$ and $\bm p_{tj}$ are in a drawing order. In addition, we make relative PE trainable to encourage DC-gra2seq to flexibly capture the generalized unseen contextual information among patches during training. And for a sketch represented by $M$ sketch patches, totally $M$ vectorized relative PEs are required in DC-gra2seq.

We simultaneously employ two types of PEs to juice both the obtainable sequential positions and the hidden contextual relationships between sketch patches from drawing orders. And PEs could be easily equipped on graph nodes and kept away from the construction of graph edges during DC-gra2seq training.

\subsection{Context-aware graphic sketch representation}

For the $M$ sketch patches cropped from the full sketch $\bm S_t$, a CNN encoder $q_{\bm\phi}(\bm v|\bm p)$, which contains seven convolutional layers (channel numbers as 8, 32, 64, 128, 256, 512 and 512) with $2\times 2$ kernels and the ReLU activation function followed by max pooling and batch normalization, is employed to capture the patch embeddings $\bm v_{tm}\in\mathbb{R}^{512\times 1}$ of $\bm p_{tm}$. $\bm v_{tm}$, regarded as a graph node, shares the same length with $\bm P$ and $\bm R$.

We link sketch patches $\bm p_{ti}$ and $\bm p_{tj}$ to construct a graph edge by considering their synonymous proximity \cite{zang2023linking}, computed by the cosine similarity between their captured patch embeddings $\bm v_{ti}$ and $\bm v_{tj}$.
\begin{equation}
	\label{eq:synonymous_proximity}
	\alpha_t(i,j)=\cos(\bm v_{ti}, \bm v_{tj})=\frac{\bm v_{ti}^\top\bm v_{tj}}{\Vert\bm v_{ti}\Vert_2\cdot\Vert\bm v_{tj}\Vert_2},
\end{equation}
where $\Vert\cdot\Vert_2$ denotes the L2 norm. A large value of $\alpha_t(i,j)$ indicates that the patch $\bm p_{ti}$ is highly relevant to $\bm p_{tj}$ in content. It is worth noting that neither the absolute PE nor relative PE attends the calculation of $\alpha_t(i, j)$. The absence of PEs reduces the uncertainty in node linking brought by variants of sketch drawings, driving to reliable edge constructions for the passing message aggregation.

We utilize the computed $\alpha_t(i,j)$ in Eq. (\ref{eq:synonymous_proximity}) to construct graph edges, and each node is only linked to the ones with the top-2 largest values among $\{\bm v_{tm}\}_{m\neq i}$. We arrive at the masked coefficients, stored in an adjacency matrix $\bm A_t$, with the element $\bm A_t(i,j)$ in $i$-th row and $j$-th column computed by,
\begin{equation}
	\label{eq:element_a}
	\bm A_t(i,j)=\left\{
			\begin{array}{ll}
				1, & j=i,\\
        		0.5\cdot \alpha_t(i,j), & j=j^*, \\
				& j^*=\mathop{\arg\max}\limits_{m\neq i}~\alpha_t(i,m),\\
        		0.2\cdot \alpha_t(i,j), & j=j',\\
				& j'=\mathop{\arg\max}\limits_{m\neq i, j^*}~\alpha_t(i,m),\\
        		0, & \text{otherwise.}
    			\end{array}\right.
\end{equation}
Following the settings in \cite{zang2023linking}, we offer the top-2 neighbors with the weights 0.5 and 0.2, according to their importance to allow the more relevant neighbor to transport more beneficial messages towards the target node during the aggregation process by the following GCN encoder.

With the constructed graph edges stored in $\bm A_t$, we activate the message aggregating process for learning node $\bm v_{ti}$ via a GCN layer, shown in Eq. (\ref{eq:aggregation_a_node}).
\begin{align}
	\label{eq:aggregation_a_node}
	\bm v^{\text{aggr}}_{ti}=\sum_{\bm v_{tj}\in\{\mathcal{N}(\bm v_{ti}),\bm v_{ti}\}}\Big[\bm A_t(i, j)\big(\bm v_{tj}+\bm R(i,j)\big)\Big]+\bm P(i),
\end{align}
where $\mathcal{N}(\bm v_{ti})$ denotes the neighborhood of $\bm v_{ti}$ determined by the adjacency matrix $\bm A_t$. 

As the adjacency matrix $\bm A_t$ is sparse, a target node $\bm v_{ti}$ only receives the messages passing from two neighboring nodes, whose sketch patches are the most relevant in semantics. The neighboring node $\bm v_{tj}$ delivers not only its captured patch features $\bm v_{tj}$ but also the corresponding contextual relationships embedded by $\bm R(i, j)$. It encourages DC-gra2seq to realize what semantically related features a neighboring node has, and how far away the node is toward $\bm v_{ti}$ along the drawing order, equipping local patch patterns with contextual information observed from a global view. In addition, after incorporating all the passing messages, we provide each node with its sequential positions in the drawing order by $\bm P(i)$, to further enrich the visual patterns with temporal information from the drawing orders. It is worth noting that both absolute and relative PEs only participate in node aggregation, but are kept away from the edge construction in $\bm A_t$, driving to variant-drawing-protected sketch representation.

In practice, a GCN encoder $q_{\bm\xi,\bm R}(\bm y|\bm V,\bm A)$ is introduced to incorporate all patch embeddings along with their absolute and relative PEs. Furthermore, a full sketch embedding $\bm v_{t0}$, which is also captured by the CNN encoder $q_{\bm\phi}$ from a $256\times 256$ image $\bm p_{t0}$ resized from the original full sketch $\bm S_t$, is also taken into account during message aggregation. Eq. (\ref{eq:aggregation}) presents the GCN layer.
\begin{align}
	\bm\mu_t, \bm\sigma_t&=\text{MLP}\left[\tilde{\bm D}_t^{-\frac{1}{2}}\tilde{\bm A}_t\tilde{\bm D}_t^{-\frac{1}{2}}(\tilde{\bm V}^\top_t+\tilde{\bm R}^\top)+\tilde{\bm P}^\top\right],\label{eq:aggregation}\\
	\tilde{\bm V}_t&=\left[\bm v_{t0}, \bm V_t\right]=\left[\bm v_{t0}, \bm v_{t1}, \cdots, \bm v_{tM}\right],\nonumber\\
	\tilde{\bm A}_t&=\left[
		\begin{array}{cc}
			0.5 & \bm 0^\top\\
			0.5\cdot\bm 1 & \bm A_t
		\end{array}\right],
		\tilde{\bm P}=\left[\bm\delta_0, \bm P\right],
		\tilde{\bm R}=\left[\bm\delta'_0, \bm R\right],
	\label{eq:stack}
\end{align}
where $\tilde{\bm D}_t$ is the degree matrix of $\tilde{\bm A}_t$. $\tilde{\bm A}_t$, $\tilde{\bm V}_t$, $\tilde{\bm R}$ and $\tilde{\bm P}$, which are defined in Eq. (\ref{eq:stack}), contain static placeholders ($\bm 0$, $\bm 1$, $\bm\delta_0$, $\bm\delta'_0\in\mathbb{R}^{M\times 1}$) to fit the matrix calculation with the newly introduced embedding $\bm v_{t0}$. These placeholders are vectors with fixed values which cannot be trained. We incorporate the features extracted in a global view from the full sketch with the ones captured from sketch patches with local sketch details for the final sketch representation.

$\text{MLP}(\cdot)$ in Eq. (\ref{eq:aggregation}) stands for a multi-layer perception, which produces the vectors $\bm\mu_t$ and $\bm\sigma_t$ to compute the final sketch representation $\bm y_t=\bm\mu_t+\bm\sigma_t\odot\bm\epsilon$ via reparameterization \cite{kingma2013auto}. $\bm\epsilon$ is randomly sampled from the standard Gaussian distribution $\mathcal{G}(\bm\epsilon|\bm 0, \bm I)$.

\subsection{Training a DC-gra2seq}

We feed the captured sketch representation $\bm y_t$ into an RNN decoder $p_{\bm\theta}(\bm S|\bm y)$, which is adopted from \cite{ha2017neural}, to reconstruct the input sketch in a sequential format. Our objective is to maximize the log-likelihood term,
\begin{align}
	\label{eq:loss}
	\mathcal{L}(\bm\theta, \bm\phi, \bm\xi, \bm R|\bm S)=\sum_{t=1}^N\text{E}_{q_{\bm\phi,\bm\xi,\bm R}(\bm y_t|\bm S_t)}\log p_{\bm\theta}(\bm S_t|\bm y_t).
\end{align}
Though DC-gra2seq falls into the variational auto-encoder (VAE) \cite{kingma2013auto} framework, we follow \cite{su2020sketchhealer,qi2022generative,zang2023linking,chen2017sketch} to remove the Kullback-Leibler (KL) divergence term from the objective to learn sketch representation more freely.

\section{Experiments and analysis}

We testify our DC-gra2seq on two tasks, controllable sketch synthesis \cite{zang2021controllable} and sketch healing \cite{su2020sketchhealer}, aiming to verify whether DC-gra2seq learns accurate and robust sketch representations.

\subsection{Preparations}

\textbf{Datasets}. Three datasets from QuickDraw \cite{ha2017neural} are selected, and all of them are adopted from \cite{zang2023linking} to make a direct comparison with SP-gra2seq \cite{zang2023linking}. Dataset 1 (DS1), which contains bee, bus, flower, giraffe and pig, evaluates representing performance on large variations of sketches within the same category. Dataset 2 (DS2), which collects airplane, angel, apple, butterfly, bus, cake, fish, spider, the Great Wall and umbrella, evaluates whether the models are sensitive to categorical patterns. Dataset 3 (DS3) collects the five categories in DS1 along with three new ones (car, cat and horse). Sketches in different categories are with shared stylistic patterns, e.g., both giraffes and horses can head left or right. Each category contains $70K$ training, $2.5K$ valid and $2.5K$ test sketches ($1K=1000$).

\textbf{Baselines}. We compare DC-gra2seq with eleven baseline models. Song et al. \cite{song2018learning} and Xu et al. \cite{xu2020learning} learn representations from sketch images and their corresponding sketch sequences simultaneously. Sketch-pix2seq \cite{chen2017sketch} and Lmser-pix2seq \cite{li2024lmser} learn representations from sketch images. We make them a group to verify the advantage brought by drawing orders as an additional source. RPCL-pix2seq \cite{zang2021controllable} and RPCL-pix2seqH \cite{zang2024self} both constrain sketch codes with a specific distribution. SketchLattice \cite{qi2021sketchlattice}, SP-gra2seq \cite{zang2023linking}, SketchHealer \cite{su2020sketchhealer} and SketchHealer 2.0 \cite{qi2022generative} learn graphic sketch representations. The sequential information from drawing orders are not used in SketchLattice and SP-gra2seq, while they are treated as prior knowledge to construct graph edges in SketchHealer and SketchHealer 2.0. Moreover, following \cite{zang2023linking}, we provide SketchLattice$^+$ to replace the original node embeddings captured from coordinates on lattice with the ones from sketch patches, to make a fair comparison.

\textbf{Metrics}. We use $Rec$ and $Ret$ \cite{zang2021controllable} for evaluation. $Rec$ describes whether a generated sketch $\hat{\bm S}_t$ and its corresponding input $\bm S_t$ are in the same category. We pre-train three sketch-a-net \cite{yu2015sketch} classifiers to compute $Rec$ for three datasets, respectively. $Ret$ indicates whether $\hat{\bm S}_t$ is successfully controlled by preserving both categorical and stylistic patterns from $\bm S_t$. When calculating $Ret$ for $\bm S_t$, we obtain its code $\bm y_t$ and the reconstruction $\hat{\bm S}_t$. We retrieve $\bm y_t$ from the gallery $\bm Y=\{\bm y_t(\bm S_t)|\bm S_t\in\text{test set}\}$ with the code $\hat{\bm y}_t$ of $\hat{\bm S}_t$, and $Ret$ indicates the successful retrieving rate. Both $Rec$ and $Ret$ are calculated from the entire test set.

When training a DC-gra2seq, the patch number $M$ and the mini-batch size $N$ are fixed at $20$ and $256$, respectively. We adopt the Adam optimizer to learn DC-gra2seq, and the learning rate starts from $10^{-3}$ with a decay rate of $0.95$ after each training epoch.

\subsection{Controllable sketch synthesis}

Controllable sketch synthesis requires a model to generate sketches with the expected categorical and stylistic patterns. Table \ref{tab:controllable_synthesis_ds1}-\ref{tab:controllable_synthesis_ds3} report the quantitative performances on three datasets, respectively.

\begin{table}[!h]
	\centering
	\caption{Controllable sketch synthesis performance (\%) on DS1. ``@$k$'' indicates the top-$k$ accuracy.}
	\label{tab:controllable_synthesis_ds1}
	\begin{tabular}{lcrrr}
		\hline
		\multirow{2}*{Model} & \multirow{2}*{$Rec\uparrow$} & \multicolumn{3}{c}{$Ret\uparrow$} \\ \cline{3-5}
		~ & ~ & @1 & @10 & @50\\
		\hline
		sketch-pix2seq \cite{chen2017sketch} & 83.99 & 13.45 & 30.12 & 49.99 \\
		Song et al. \cite{song2018learning} & 91.77 & 16.41 & 36.43 & 52.22 \\
		Xu et al. \cite{xu2020learning} & 93.32 & 22.70 & 54.40 & 75.04 \\
		RPCL-pix2seq \cite{zang2021controllable} & 93.18 & 17.86 & 38.87 & 55.30\\
		RPCL-pix2seqH \cite{zang2024self} & 94.97 & 71.36 & 90.29 & 94.86 \\
		SketchHealer \cite{su2020sketchhealer} & 91.04 & 58.80 & 82.15 & 91.33 \\
		SketchHealer 2.0 \cite{qi2022generative} & 93.13 & 57.19 & 84.54 & 90.26 \\
		SketchLattice \cite{qi2021sketchlattice} & 75.91 & 6.55 & 14.01 & 26.72 \\
		SketchLattice$^+$ & 95.18 & 72.74 & 91.60 & 97.14 \\
		Lmser-pix2seq \cite{li2024lmser} & 95.26 & 92.90 & 97.72 & 99.02 \\
		SP-gra2seq \cite{zang2023linking} & 95.91 & 94.88 & 99.11 & 99.72 \\
		DC-gra2seq & \pmb{96.26} & \pmb{96.84} & \pmb{99.48} & \pmb{99.88} \\
		\hline
	\end{tabular}
\end{table}

\begin{table}[!h]
	\centering
	\caption{Controllable sketch synthesis performance (\%) on DS2.}
	\label{tab:controllable_synthesis_ds2}
	\begin{tabular}{lcrrr}
		\hline
		\multirow{2}*{Model} & \multirow{2}*{$Rec\uparrow$} & \multicolumn{3}{c}{$Ret\uparrow$} \\ \cline{3-5} 
		~ & ~ & @1 & @10 & @50 \\
		\hline
		sketch-pix2seq \cite{chen2017sketch} & 85.46 & 50.94 & 71.38 & 80.15 \\
		Song et al. \cite{song2018learning} & 86.98 & 58.84 & 76.84 & 80.06 \\
		Xu et al. \cite{xu2020learning} & 90.01 & 51.98 & 75.79 & 83.26 \\
		RPCL-pix2seq \cite{zang2021controllable} & 88.73 & 53.19 & 71.60 & 87.91 \\
		RPCL-pix2seqH \cite{zang2024self} & 92.62 & 85.91 & 93.67 & 95.90 \\
		SketchHealer \cite{su2020sketchhealer} & 94.04 & 87.54 & 96.19 & 98.26 \\
		SketchHealer 2.0 \cite{qi2022generative} & 90.94 & 87.37 & 94.59 & 97.60 \\
		SketchLattice \cite{qi2021sketchlattice} & 71.80 & 6.91 & 14.76 & 28.82 \\
		SketchLattice$^+$ & 94.30 & 90.56 & 97.78 & 99.27 \\
		Lmser-pix2seq \cite{li2024lmser} & 94.27 & 89.95 & 97.08 & 99.25 \\
		SP-gra2seq \cite{zang2023linking} & 94.85 & 90.83 & \pmb{98.29} & 99.08 \\
		DC-gra2seq & \pmb{95.70} & \pmb{92.75} & 98.08 & \pmb{99.42} \\
		\hline
	\end{tabular}
\end{table}

\begin{table}[!h]
	\centering
	\caption{Controllable sketch synthesis performance (\%) on DS3.}
	\label{tab:controllable_synthesis_ds3}
	\begin{tabular}{lcrrr}
		\hline
		\multirow{2}*{Model} & \multirow{2}*{$Rec\uparrow$} & \multicolumn{3}{c}{$Ret\uparrow$} \\ \cline{3-5}
		~ & ~ & @1 & @10 & @50 \\
		\hline
		sketch-pix2seq \cite{chen2017sketch} & 79.13 & 22.92 & 47.55 & 58.19\\
		Song et al. \cite{song2018learning} & 83.28 & 25.47 & 43.39 & 56.16\\
		Xu et al. \cite{xu2020learning} & 82.34 & 25.25 & 54.04 & 72.90\\
		RPCL-pix2seq \cite{zang2021controllable} & 81.80 & 28.80 & 59.05 & 77.52\\
		RPCL-pix2seqH \cite{zang2024self} & 87.82 & 81.22 & 92.15 & 94.90\\
		SketchHealer \cite{su2020sketchhealer} & 87.03 & 68.52 & 82.37 & 86.57\\
		SketchHealer 2.0 \cite{qi2022generative} & 87.37 & 50.67 & 76.11 & 82.42\\
		SketchLattice \cite{qi2021sketchlattice} & 62.21 & 5.90 & 10.36 & 19.39\\
		SketchLattice$^+$ & 89.49 & 87.27 & 96.82 & 98.98\\
		Lmser-pix2seq \cite{li2024lmser} & 89.03 & 89.21 & 97.63 & 99.12 \\
		SP-gra2seq \cite{zang2023linking} & 89.83 & 94.05 & 98.72 & 99.57\\
		DC-gra2seq & \pmb{90.41} & \pmb{96.27} & \pmb{99.47} & \pmb{99.81}\\
		\hline
	\end{tabular}
\end{table}

Song et al. and Xu et al. outperform sketch-pix2seq as their sketch representations are enhanced by additional information from drawing orders. For graphic representation learning models, though SketchHealer and SketchHealer 2.0 are equipped with sketch drawing orders, they are defeated by SketchLattice$^+$ and SP-gra2seq, which never utilize the sequential information as assistance. It is because the variations of sketch drawings drive to unreliable edge constructions, and the passing messages along these edges could dilute or interrupt the target patch learning.

The proposed DC-gra2seq injects drawing orders into context-aware PEs. The sequential positions of sketch patches are embedded on graph nodes, but never participate in valuing the relationships between patches, i.e., edge construction. As a result, the position-embedded PEs provide the captured patch embeddings with contextual information, which cannot be extracted from the canvas, guiding DC-gra2seq to realize when each patch is drawn along the drawing order. In the meantime, DC-gra2seq links patches by patch embeddings only, which stabilizing the message passing between patches with synonymous contents. Thus, DC-gra2seq achieves the best controllable synthesis performance, especially on top-1 (@1) $Ret$.

\begin{figure}[t]
	\centering
	\includegraphics[width=\linewidth]{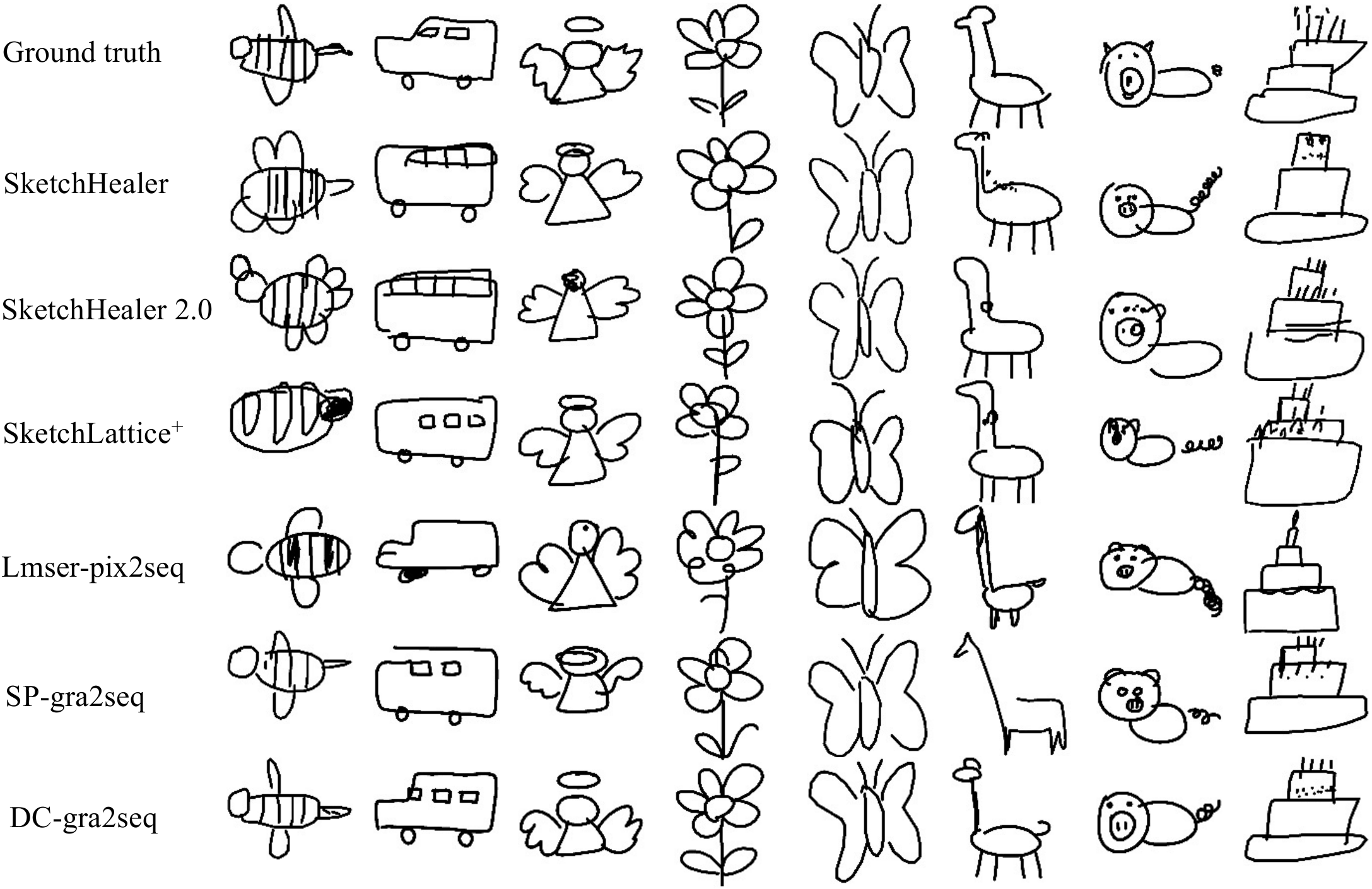}
	\caption{Qualitative comparisons on controllable sketch synthesis.}
	\label{fig:controllable_synthesis}
\end{figure}

Moreover, Fig. \ref{fig:controllable_synthesis} also presents the qualitative comparisons on controllable sketch synthesis. SketchHealer, SketchHealer 2.0, SketchLattice$^+$, Lmser-pix2seq and SP-gra2seq are selected as baselines, as they achieve high performances in quantitative comparisons. The sketches generated by DC-gra2seq could preserve more detailed patterns from inputs as the ground truth, e.g., the shape of the bus (in 2-nd column) and the candles on the cake (in 8-th column).

\begin{figure*}[h]
	\centering
	\includegraphics[width=0.9\linewidth]{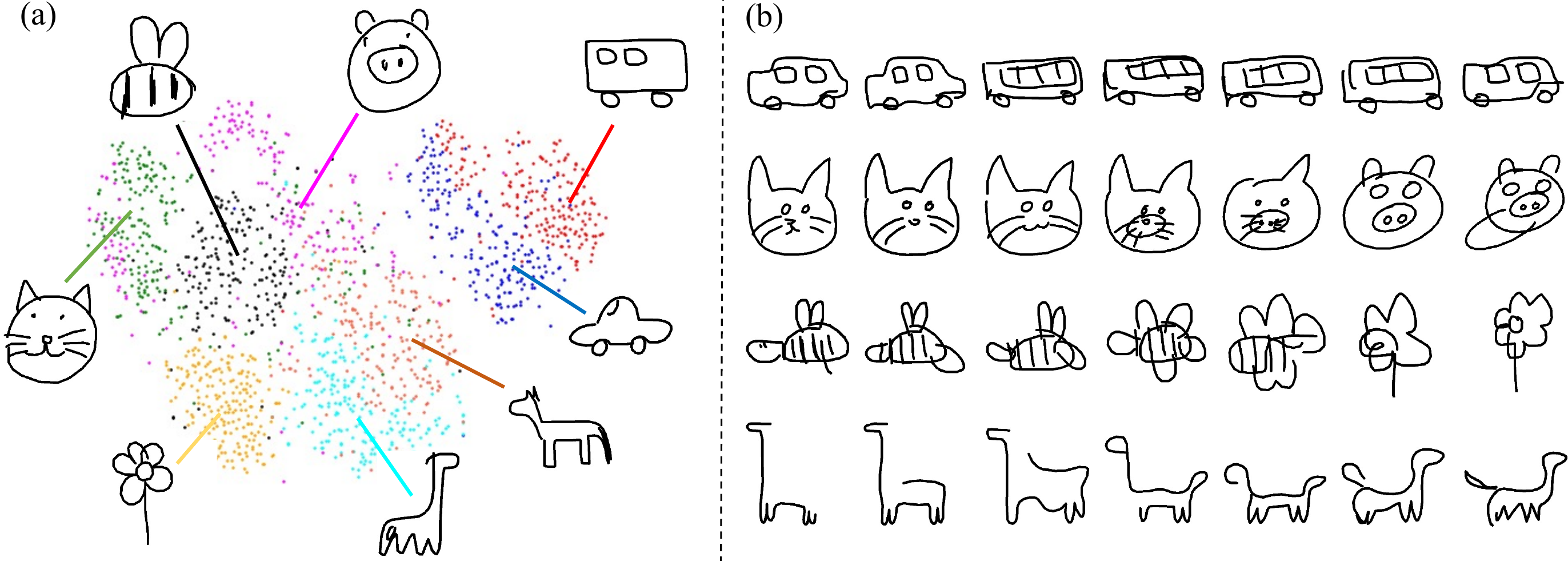}
	\caption{Generating sketches by DC-gra2seq with interpolated latent codes. (a) The latent space learned by DC-gra2seq on DS3. (b) In each row, we morph sketch patterns by interpolating their corresponding latent codes.}
	\label{fig:interpolation}
\end{figure*}

We also reveal the latent space learned by DC-gra2seq on DS3, shown in Fig. \ref{fig:interpolation}(a). Each dot indicates a captured latent code of a sketch and its color denotes the corresponding category. The high-dimensional codes are mapped into a two-dimensional space via t-SNE \cite{maaten2008visualizing}. We can figure out that the sketches in the same category are clustered in a specific latent region, making categorical patterns easier to be controlled in sketch generating process. Besides, Fig. \ref{fig:interpolation}(b) presents sketches generated by interpolated latent codes. More specifically, in each row, the latent codes are computed via a linear interpolation between the ones at two ends. The characteristics of sketches are gradually changed along with the movement of latent codes from one end to the other. It indicates that sketch patterns are positioned smoothly in latent space, which enables DC-gra2seq to flexibly control sketch patterns.

\begin{table}[!h]
	\centering
	\caption{Sketch healing performance (\%) on DS1. ``Mask'' denotes the probability for creating masks.}
	\label{tab:sketch_healing_ds1}
	\begin{tabular}{clcrrr}
		\hline
		\multirow{2}*{Mask} & \multirow{2}*{Model} & \multirow{2}*{$Rec\uparrow$} & \multicolumn{3}{c}{$Ret\uparrow$} \\ \cline{4-6}
		~ & ~ & ~ & @1 & @10 & @50 \\
		\hline
		\multirow{7}*{10\%} & SketchHealer \cite{su2020sketchhealer} & 70.38 & 14.25 & 27.91 & 45.51 \\
		~ & SketchHealer 2.0 \cite{qi2022generative} & 90.18 & 19.45 & 41.51 & 59.94 \\
		~ & SketchLattice \cite{qi2021sketchlattice} & 54.18 & 1.09 & 5.48 & 14.00 \\
		~ & SketchLattice$^+$ & 89.98 & 19.86 & 43.59 & 63.70 \\
		~ & Lmser-pix2seq \cite{li2024lmser} & 94.08 & 34.96 & 60.18 & 77.42 \\
		~ & SP-gra2seq \cite{zang2023linking} & 92.90 & 41.24 & 65.74 & 80.16 \\
		~ & DC-gra2seq & \pmb{94.45} & \pmb{74.59} & \pmb{88.83} & \pmb{94.31}  \\
		\cdashline{1-6}[2pt/2pt]
		\multirow{7}*{30\%} & SketchHealer \cite{su2020sketchhealer} & 59.00 & 0.23 & 3.48 & 10.76 \\
		~ & SketchHealer 2.0 \cite{qi2022generative} & 79.05 & 4.66 & 14.54 & 28.44 \\
		~ & SketchLattice \cite{qi2021sketchlattice} & 32.03 & 0.08 & 2.05 & 5.96 \\
		~ & SketchLattice$^+$ & 70.91 & 1.02 & 5.26 & 14.28 \\
		~ & Lmser-pix2seq \cite{li2024lmser} & 83.12 & 9.10 & 27.18 & 43.27\\
		~ & SP-gra2seq \cite{zang2023linking} & 84.85 & 12.87 & 29.39 & 45.58 \\
		~ & DC-gra2seq & \pmb{86.50} & \pmb{30.30} & \pmb{51.38} & \pmb{65.84} \\
		\hline
	\end{tabular}
\end{table}

\begin{table}[!ht]
	\centering
	\caption{Sketch healing performance (\%) on DS2.}
	\label{tab:sketch_healing_ds2}
	\begin{tabular}{clcrrr}
		\hline
		\multirow{2}*{Mask} & \multirow{2}*{Model} & \multirow{2}*{$Rec\uparrow$} & \multicolumn{3}{c}{$Ret\uparrow$} \\ \cline{4-6}
		~ & ~ & ~ & @1 & @10 & @50 \\
		\hline
		\multirow{7}*{10\%} & SketchHealer \cite{su2020sketchhealer} & 70.56 & 35.22 & 55.86 & 66.24 \\
		~ & SketchHealer 2.0 \cite{qi2022generative} & 87.07 & 39.40 & 64.01 & 81.20 \\
		~ & SketchLattice \cite{qi2021sketchlattice} & 52.97 & 1.34 & 6.86 & 17.41 \\
		~ & SketchLattice$^+$ & 90.52 & 46.98 & 70.21 & 83.12 \\
		~ & Lmser-pix2seq \cite{li2024lmser} & 92.20 & 53.99 & 73.76 & 83.61 \\
		~ & SP-gra2seq \cite{zang2023linking} & 91.24 & 50.42 & 73.35 & 85.18 \\
		~ & DC-gra2seq & \pmb{92.64} & \pmb{67.61} & \pmb{84.31} & \pmb{91.17} \\
		\cdashline{1-6}[2pt/2pt]
		\multirow{7}*{30\%} & SketchHealer \cite{su2020sketchhealer} & 61.26 & 7.98 & 19.04 & 35.03 \\
		~ & SketchHealer 2.0 \cite{qi2022generative} & 75.08 & 10.05 & 24.51 & 39.57 \\
		~ & SketchLattice \cite{qi2021sketchlattice} & 31.73 & 0.88 & 3.64 & 9.25\\
		~ & SketchLattice$^+$ & 81.76 & 10.43 & 26.44 & 42.90 \\
		~ & Lmser-pix2seq \cite{li2024lmser} & 82.00 & 10.82 & 29.02 & 48.34 \\
		~ & SP-gra2seq \cite{zang2023linking} & 82.85 & 12.19 & 29.37 & 46.53 \\
		~ & DC-gra2seq & \pmb{84.62} & \pmb{29.92} & \pmb{49.26} & \pmb{62.44} \\
		\hline
	\end{tabular}
\end{table}

\begin{table}[!ht]
	\centering
	\caption{Sketch healing performance (\%) on DS3.}
	\label{tab:sketch_healing_ds3}
	\begin{tabular}{clcrrr}
		\hline
		\multirow{2}*{Mask} & \multirow{2}*{Model} & \multirow{2}*{$Rec\uparrow$} & \multicolumn{3}{c}{$Ret\uparrow$} \\ \cline{4-6}
		~ & ~ & ~ & @1 & @10 & @50 \\
		\hline
		\multirow{7}*{10\%} & SketchHealer \cite{su2020sketchhealer} & 60.91 & 15.10 & 37.89 & 53.10\\
		~ & SketchHealer 2.0 \cite{qi2022generative} & 76.45 & 15.55 & 39.99 & 61.67\\
		~ & SketchLattice \cite{qi2021sketchlattice} & 44.14 & 0.71 & 4.19 & 11.36\\
		~ & SketchLattice$^+$ & 78.28 & 22.56 & 44.89 & 62.11\\
		~ & Lmser-pix2seq \cite{li2024lmser} & 83.75 & 35.99 & 52.31 & 69.85\\
		~ & SP-gra2seq \cite{zang2023linking} & 83.38 & 40.20 & 63.40 & 77.52\\
		~ & DC-gra2seq & \pmb{85.21} & \pmb{71.66} & \pmb{86.34} & \pmb{92.35} \\
		\cdashline{1-6}[2pt/2pt]
		\multirow{7}*{30\%} & SketchHealer \cite{su2020sketchhealer} & 48.90 & 0.43 & 7.36 & 15.79\\
		~ & SketchHealer 2.0 \cite{qi2022generative} & 60.74 & 3.75 & 11.66 & 22.36\\
		~ & SketchLattice \cite{qi2021sketchlattice} & 23.06 & 0.41 & 2.41 & 5.96\\
		~ & SketchLattice$^+$ & 67.31 & 2.75 & 11.13 & 23.68\\
		~ & Lmser-pix2seq \cite{li2024lmser} & 68.45 & 4.03 & 13.85 & 25.32 \\
		~ & SP-gra2seq \cite{zang2023linking} & 71.07 & 5.65 & 17.40 & 32.90\\
		~ & DC-gra2seq & \pmb{72.84} & \pmb{27.64} & \pmb{47.34} & \pmb{60.61}\\
		\hline
	\end{tabular}
\end{table}

\subsection{Sketch healing}

Sketch healing \cite{su2020sketchhealer} requires a model to recreate a full sketch $\hat{\bm S}_t$ from the corrupted sketch $\bm S^\text{m}_t$ with masks. The generated $\hat{\bm S}_t$ is expected to preserve both categorical and detailed stylistic characteristics from the original unmasked $\bm S_t$ ($\bm S^\text{m}_t$ is corrupted from $\bm S_t$). We adopt the approach utilized in \cite{zang2023linking} to mask sketches: After selecting the cropping centers for sketch patches, each center is applied with a masking probability of 10\% or 30\% for positioning a $256\times 256$ mask on that location on the canvas. For SketchLattice and SketchLattice$^+$, the coordinate selection on lattice is proceeded after applying masks.

We still evaluate sketch healing by $Rec$ and $Ret$ following \cite{zang2023linking}, but adjust $Ret$ to fit this task. Firstly, we mask $\bm S_t$ to get the corrupted sketch input $\bm S^\text{m}_t$, which is sent into a model to generate $\hat{\bm S}_t$. Then we compute the codes $\bm y_t$ and $\hat{\bm y}_t$ for $\bm S_t$ and $\hat{\bm S}_t$, respectively. Finally, we retrieve the exact $\bm y_t$ from $\bm Y=\{\bm y_t(\bm S_t)|\bm S_t\in\text{test set}\}$ with $\hat{\bm y}_t$ to compute $Ret$. It is worth noting that each sketch is banded with the same masks during the metrics calculation for different models to make the comparison fair.

Table \ref{tab:sketch_healing_ds1}-\ref{tab:sketch_healing_ds3} present the sketch healing results on three datasets, respectively. Our DC-gra2seq yields significant improvements on $Ret$. When dealing with sketch patches corrupted by masks, the passing messages from the semantic-related patches by the values of cosine similarity could contain valuable missing information under the masks. Moreover, by taking account of PEs, DC-gra2seq might realize where these semantic-related patches are drawn from the contextual positions in a drawing order. By incorporating both visual patterns from the canvas and their sequential information from the embedded PEs, DC-gra2seq learns much more comprehensive and accurate sketch representations.

\begin{figure*}[t]
	\centering
	\includegraphics[width=0.9\linewidth]{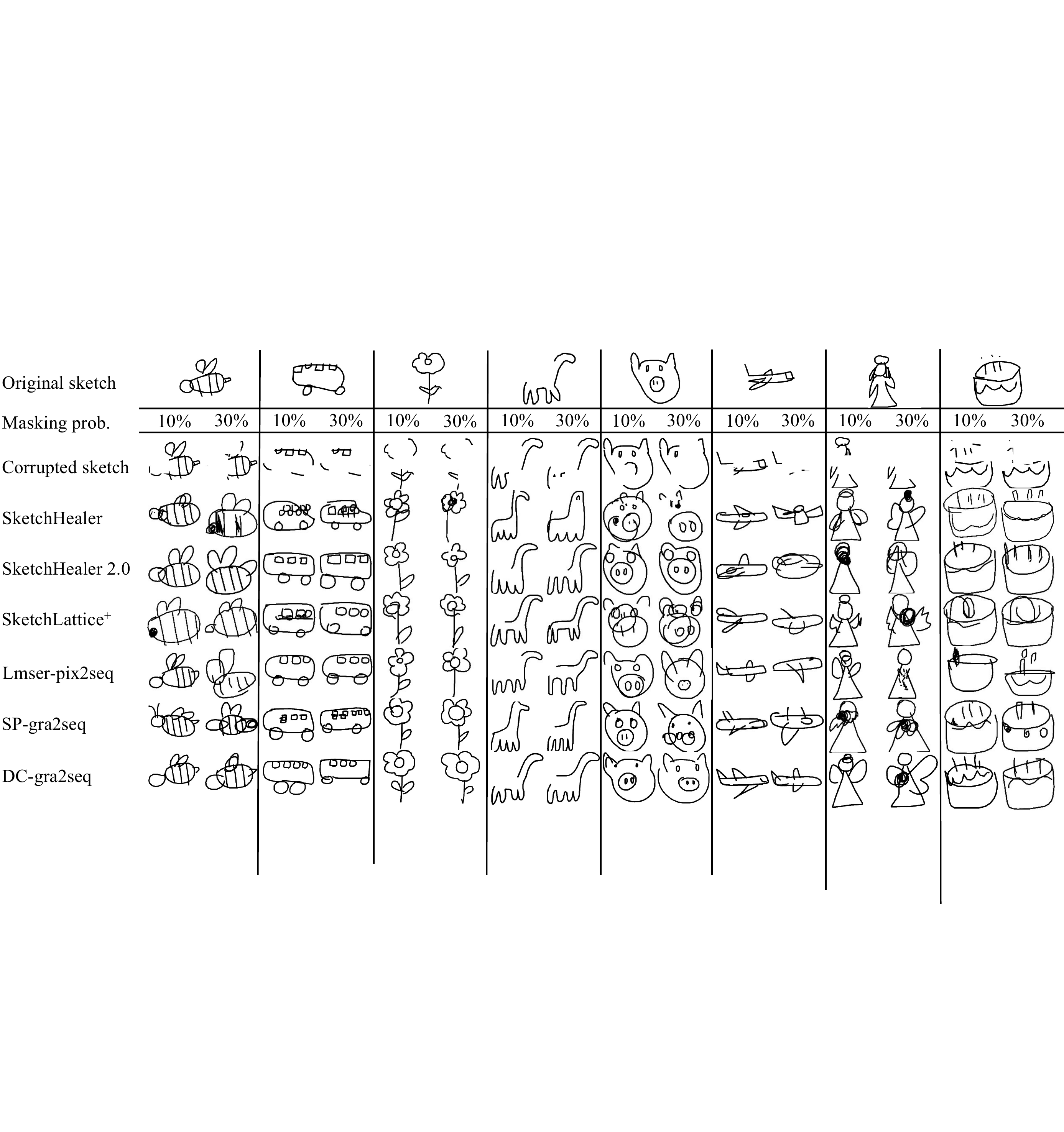}
	\caption{Qualitative comparisons on sketch healing.}
	\label{fig:sketch_healing}
\end{figure*}

We also present some sketch healing results in Fig. \ref{fig:sketch_healing}, as qualitative comparisons. When facing a masking probability of 30\%, some key characteristics of sketches are corrupted, e.g., a pig with missing nose. By making use of drawing orders, DC-gra2seq could access where these masked patches are drawn in the drawing order, and collect messages from contextual patches to generate a recognizable pig from the remaining face counter.

\subsection{Variant-drawing-protected DC-gra2seq}

This section aims to verify whether our DC-gra2seq could reduce the impact brought by variants of sketch drawings, to further achieve robust sketch representations.

\begin{figure*}[!ht]
	\centering
	\includegraphics[width=0.8\linewidth]{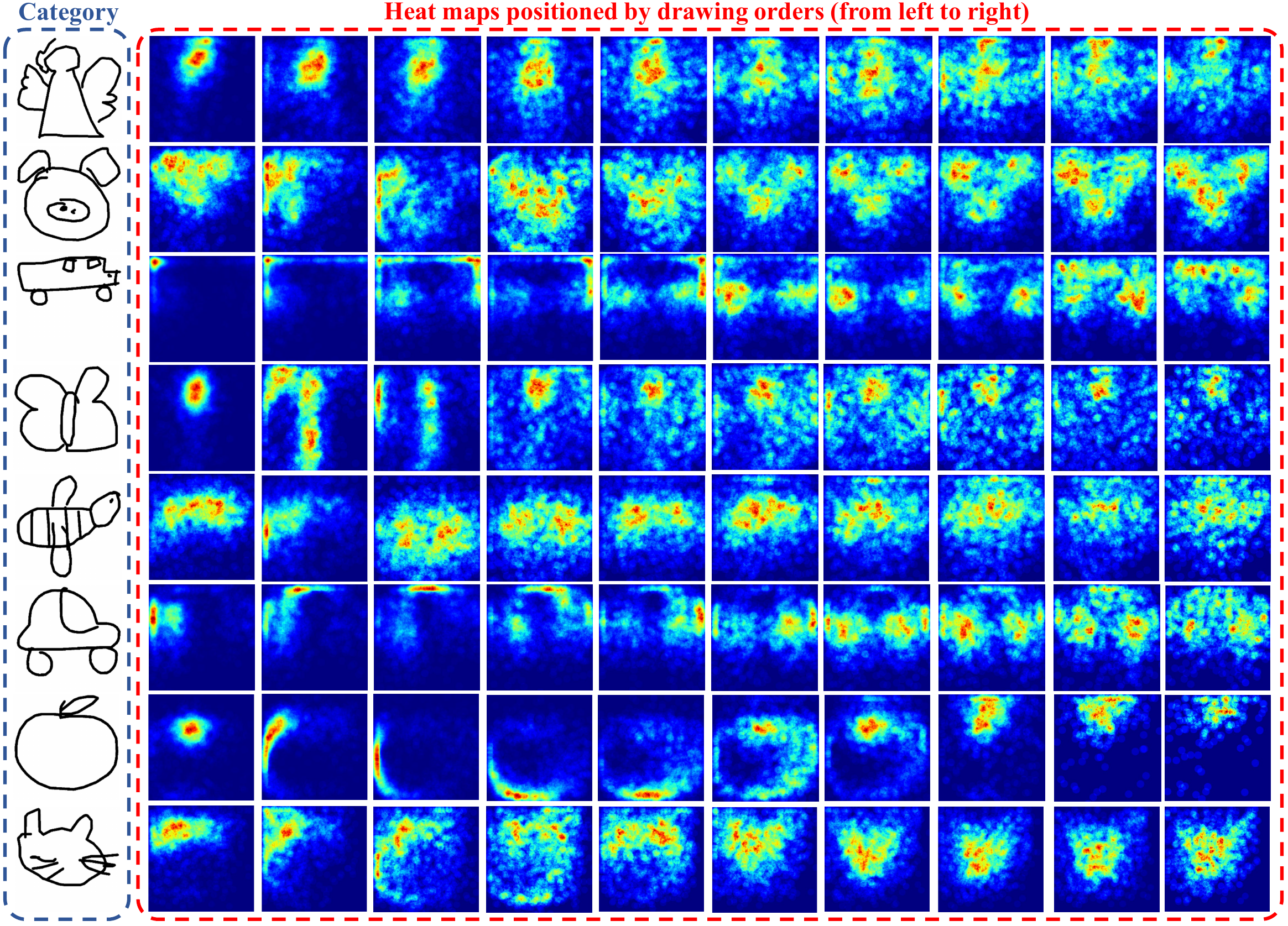}
	\caption{Variations of sketch drawings revealed by spatial heat maps on eight exemplary categories (angel, pig, bus, butterfly, bee, car, apple and cat), respectively. In each row, the heat maps from left to right are organized by sketch drawing orders. A heat map reports the magnitude where the next sketch component would be drawn on the canvas, and the red regions indicate the positions with a large probability to meet the next stroke at that moment.}
	\label{fig:sketch_drawing}
\end{figure*}

\begin{table*}[!ht]
	\centering
	\caption{Controllable sketch synthesis performance (\%) respectively computed on every single category in DS1. By checking each column, we figure out that pigs contribute the least to the performance on the full DS1, due to serious variants of sketch drawings revealed by Fig. \ref{fig:sketch_drawing}.}
	\label{tab:protection}
	\begin{tabular}{@{}lcrrrcrrrcrrr@{}}
		\hline
		\multirow{3}*{Category} & \multicolumn{4}{c}{SketchHealer 2.0} & \multicolumn{4}{c}{SP-gra2seq} & \multicolumn{4}{c}{DC-gra2seq}\\ \cline{2-13}
		~ & \multirow{2}*{$Rec$} & \multicolumn{3}{c}{$Ret$} & \multirow{2}*{$Rec$} & \multicolumn{3}{c}{$Ret$} & \multirow{2}*{$Rec$} & \multicolumn{3}{c}{$Ret$}\\ 
		\cline{3-5} \cline{7-9} \cline{11-13}
		~ & ~ & @1 & @10 & @50 & ~ & @1 & @10 & @50 & ~ & @1 & @10 & @50\\
		\hline
		Bee & 89.52 & 30.72 & 57.60 & 75.72 & 93.48 & 92.16 & 96.36 & 98.92 & 94.20 & 97.44 & 99.64 & 99.96\\
		Bus & 98.60 & 57.44 & 83.88 & 94.24 & 98.88 & 97.00 & 99.38 & 99.89 & 99.00 & 97.12 & 99.64 & 100.00\\
		Flower & 94.52 & 33.16 & 62.08 & 79.56 & 96.76 & 91.28 & 94.16 & 97.32 & 98.36 & 96.80 & 99.12 & 99.72\\
		Giraffe & 94.48 & 41.96 & 75.20 & 91.96 & 96.16 & 87.52 & 98.94 & 99.76 & 96.84 & 97.48 & 99.88 & 100.00\\
		Pig & 89.20 & 24.76 & 51.76 & 73.20 & 92.88 & 83.56 & 90.68 & 96.32 & 93.72 & 96.64 & 99.76 & 99.96\\
		\cdashline{1-13}[2pt/2pt]
		Full DS1 & 93.13 & 57.19 & 84.54 & 90.26 & 95.91 & 94.88 & 99.11 & 99.72 & 96.26 & 96.84 & 99.48 & 99.88\\
		\hline
	\end{tabular}
\end{table*}

We figure out that sketches in some category suffer more serious variants of sketch drawings than the others. Fig. \ref{fig:sketch_drawing} visualizes the degree of variants by using heat maps on eight exemplary categories. Each row contains ten heat maps positioned by sketch drawing order from left to right. And each map reports the magnitude where the next sketch component will be drawn on the canvas. The red regions indicate the positions with a large probability to meet the next stroke at that moment. By reviewing the heat maps in a row from left to right, we can realize how serious the variants of drawings are for a category. For example, drawing a pig is more free, since huge areas are colored in red, yellow and green. But drawing a bus is more regular with more concentrated focuses.

The deviation of variant levels among categories inspires us to further explore the connection between the approaches for using drawing orders and their sketch representing performance. We select SketchHealer 2.0, SP-gra2seq and DC-gra2seq for comparison, since they make (no) use of drawing orders by different approaches. We compute their controllable synthesis performances on every single category in DS1, shown in Table \ref{tab:protection}. By checking the values in each column, we find that buses contribute the most while pigs contribute the least to the performance on the full DS1, which is consistent with the observation in Fig. \ref{fig:sketch_drawing}. It supports the assumption that variations of sketch drawings could reduce the representing performance as expected. Furthermore, our DC-gra2seq yields the smallest gap on performance between pig and bus, while SketchHealer 2.0 gets the largest. It could be an evidence to support our  DC-gra2seq is protected from variants of sketch drawings by injecting drawing orders into graph nodes.

\begin{table}[!h]
	\centering
	\caption{Controllable sketch synthesis (Mask~=~0\%) and sketch healing (Mask~=~10\%, 30\%) performances of DC-gra2seq on DS1, by allowing PEs to participate in the construction of graph nodes, edges or not. ``Node $\checkmark$ Edge $\times$'' denotes the proposed DC-gra2seq.}
	\label{tab:ablation_pe_ds1}
	\begin{tabular}{ccccrrr}
		\hline
		\multirow{2}*{Mask} & \multirow{2}*{Node} & \multirow{2}*{Edge} & \multirow{2}*{$Rec$} & \multicolumn{3}{c}{$Ret$} \\ \cline{5-7}
		~ & ~ & ~ & ~ & @1 & @10 & @50 \\
		\hline
		\multirow{4}*{0\%} & $\times$ & $\times$ & 94.51 & 80.90 & 95.09 & 98.19 \\
		~ & $\checkmark$ & $\times$ & \pmb{96.26} & \pmb{96.84} & \pmb{99.48} & \pmb{99.88} \\
		~ & $\times$ & $\checkmark$ & 95.69 & 79.22 & 93.92 & 97.77 \\
		~ & $\checkmark$ & $\checkmark$ & 95.91 & 95.66 & 99.23 & 99.78 \\
		\cdashline{1-7}[2pt/2pt]
		\multirow{4}*{10\%} & $\times$ & $\times$ & 92.55 & 40.69 & 65.64 & 79.69 \\
		~ & $\checkmark$ & $\times$ & \pmb{94.45} & \pmb{74.59} & \pmb{88.83} & \pmb{94.31} \\
		~ & $\times$ & $\checkmark$ & 93.48 & 50.64 & 75.10 & 87.46 \\
		~ & $\checkmark$ & $\checkmark$ & 93.96 & 70.56 & 86.10 & 92.56 \\
		\cdashline{1-7}[2pt/2pt]
		\multirow{4}*{30\%} & $\times$ & $\times$ & 82.18 & 10.93 & 26.52 & 42.02 \\
		~ & $\checkmark$ & $\times$ & \pmb{86.50} & \pmb{30.30} & \pmb{51.38} & \pmb{65.84} \\
		~ & $\times$ & $\checkmark$ & 84.23 & 16.28 & 36.80 & 55.01 \\
		~ & $\checkmark$ & $\checkmark$ & 85.00 & 26.15 & 45.94 & 61.06 \\
		\hline
	\end{tabular}
\end{table}

\begin{table}[!h]
	\centering
	\caption{Controllable sketch synthesis (Mask~=~0\%) and sketch healing (Mask~=~10\%, 30\%) performances of DC-gra2seq on DS2, by allowing PEs to participate in the construction of graph nodes, edges or not.}
	\label{tab:ablation_pe_ds2}
	\begin{tabular}{ccccrrr}
		\hline
		\multirow{2}*{Mask} & \multirow{2}*{Node} & \multirow{2}*{Edge} & \multirow{2}*{$Rec$} & \multicolumn{3}{c}{$Ret$} \\ \cline{5-7}
		~ & ~ & ~ & ~ & @1 & @10 & @50 \\
		\hline
		\multirow{4}*{0\%} & $\times$ & $\times$ & 85.84 & 41.00 & 71.80 & 87.31 \\
		~ & $\checkmark$ & $\times$ & \pmb{95.70} & \pmb{92.75} & \pmb{98.08} & \pmb{99.42} \\
		~ & $\times$ & $\checkmark$ & 92.41 & 67.49 & 88.30 & 95.30 \\
		~ & $\checkmark$ & $\checkmark$ & 94.67 & 90.00 & 97.88 & 99.32 \\
		\cdashline{1-7}[2pt/2pt]
		\multirow{4}*{10\%} & $\times$ & $\times$ & 90.08 & 41.08 & 66.21 & 80.72 \\
		~ & $\checkmark$ & $\times$ & \pmb{92.64} & \pmb{67.61} & \pmb{84.31} & \pmb{91.17} \\
		~ & $\times$ & $\checkmark$ & 89.48 & 45.24 & 70.13 & 83.33 \\
		~ & $\checkmark$ & $\checkmark$ & 91.55 & 67.10 & 84.08 & 91.02 \\
		\cdashline{1-7}[2pt/2pt]
		\multirow{4}*{30\%} & $\times$ & $\times$ & 77.20 & 10.80 & 25.61 & 45.56 \\
		~ & $\checkmark$ & $\times$ & \pmb{84.62} & \pmb{29.92} & \pmb{49.26} & \pmb{62.44} \\
		~ & $\times$ & $\checkmark$ & 80.40 & 18.62 & 37.78 & 53.70 \\
		~ & $\checkmark$ & $\checkmark$ & 83.94 & 29.38 & 48.55 & 62.22 \\
		\hline
	\end{tabular}
\end{table}

\begin{table}[!h]
	\centering
	\caption{Controllable sketch synthesis (Mask~=~0\%) and sketch healing (Mask~=~10\%, 30\%) performances of DC-gra2seq on DS3, by allowing PEs to participate in the construction of graph nodes, edges or not.}
	\label{tab:ablation_pe_ds3}
	\begin{tabular}{ccccrrr}
		\hline
		\multirow{2}*{Mask} & \multirow{2}*{Node} & \multirow{2}*{Edge} & \multirow{2}*{$Rec$} & \multicolumn{3}{c}{$Ret$} \\ \cline{5-7}
		~ & ~ & ~ & ~ & @1 & @10 & @50 \\
		\hline
		\multirow{4}*{0\%} & $\times$ & $\times$ & 83.93 & 57.50 & 83.52 & 93.45\\
		~ & $\checkmark$ & $\times$ & \pmb{90.41} & \pmb{96.27} & \pmb{99.47} & \pmb{99.81}\\
		~ & $\times$ & $\checkmark$ & 88.48 & 76.25 & 92.64 & 97.24\\
		~ & $\checkmark$ & $\checkmark$ & 89.05 & 94.74 & 99.44 & 99.75\\
		\cdashline{1-7}[2pt/2pt]
		\multirow{4}*{10\%} & $\times$ & $\times$ & 83.84 & 39.03 & 62.23 & 76.74\\
		~ & $\checkmark$ & $\times$ & \pmb{85.21} & \pmb{71.66} & 86.34 & 92.35\\
		~ & $\times$ & $\checkmark$ & 84.79 & 46.77 & 70.37 & 82.90\\
		~ & $\checkmark$ & $\checkmark$ & 85.18 & 69.96 & \pmb{86.67} & \pmb{92.43}\\
		\cdashline{1-7}[2pt/2pt]
		\multirow{4}*{30\%} & $\times$ & $\times$ & 70.35 & 4.72 & 15.05 & 29.47\\
		~ & $\checkmark$ & $\times$ & \pmb{72.84} & \pmb{27.64} & \pmb{47.34} & \pmb{60.61}\\
		~ & $\times$ & $\checkmark$ & 71.99 & 14.56 & 31.54 & 47.96\\
		~ & $\checkmark$ & $\checkmark$ & 72.26 & 25.88 & 46.33 & 59.82\\
		\hline
	\end{tabular}
\end{table}

\subsection{The impact of injecting drawing orders into nodes or edges}

We further discuss the impact of injecting sketch drawing orders into graph nodes, edges or both via PEs. To make PEs access to edge construction, we allow the absolute PEs $\bm P$ to participate in the computation of the coefficients $\alpha_t$ directly, i.e., replacing Eq. (\ref{eq:synonymous_proximity}) by $\alpha_t(i,j)=(\bm v_{ti}+\bm P(i))^\top(\bm v_{tj}+\bm P(j))$. A softmax activation function is applied over $\bm A_t$ in Eq. (\ref{eq:element_a}) before sent into Eq. (\ref{eq:aggregation}) for message aggregation. 

Table \ref{tab:ablation_pe_ds1}-\ref{tab:ablation_pe_ds3} present the experimental results on three datasets, respectively. When PEs attend the edge construction only (Node $\times$, Edge $\checkmark$), it outperforms the version which never uses PEs (Node $\times$, Edge $\times$) in most cases. The usage of drawing orders via PEs embedded graph edges is beneficial to improve sketch learning, but the performance gain is sensitive to various sketch drawings, e.g., it significantly outperforms ``Node $\times$, Edge $\times$'' on $Rec$ in DS2, but is defeated on top-1, top-10 $Ret$ in DS1. Similarly, our DC-gra2seq (Node $\checkmark$, Edge $\times$) outperforms ``Node $\checkmark$, Edge $\checkmark$'' in most cases. As a result, embedding drawing orders into graph nodes is a proper way to make better use of sequential information.

\begin{table}[!ht]
	\centering
	\caption{Comparisons between variants of DC-gra2seq on DS1, by activating two types of PEs, or not. Mask~=~0\% and Mask~=~10\%, 30\% indicate controllable sketch synthesis and sketch healing, respectively.}
	\label{tab:ablation_2pe_ds1}
	\begin{tabular}{ccccrrr}
		\hline
		\multirow{2}*{Mask} & \multirow{2}*{\makecell{Absolute\\PE}} & \multirow{2}*{\makecell{Relative\\PE}} & \multirow{2}*{$Rec$} & \multicolumn{3}{c}{$Ret$} \\ \cline{5-7}
		~ & ~ & ~ & ~ & @1 & @10 & @50 \\
		\hline
		\multirow{4}*{0\%} & $\times$ & $\times$ & 95.91 & 94.88 & 99.11 & 99.72 \\
		~ & $\checkmark$ & $\times$ & 96.14 & 95.61 & 99.39 & 99.84 \\
		~ & $\times$ & $\checkmark$ & 95.90 & 94.78 & 99.13 & 99.63 \\
		~ & $\checkmark$ & $\checkmark$ & \pmb{96.26} & \pmb{96.84} & \pmb{99.48} & \pmb{99.88} \\
		\cdashline{1-7}[2pt/2pt]
		\multirow{4}*{10\%} & $\times$ & $\times$ & 92.90 & 41.24 & 65.74 & 80.16 \\
		~ & $\checkmark$ & $\times$ & 94.17 & 72.45 & 88.38 & 93.92 \\
		~ & $\times$ & $\checkmark$ & 93.61 & 59.17 & 78.50 & 88.34 \\
		~ & $\checkmark$ & $\checkmark$ & \pmb{94.45} & \pmb{74.59} & \pmb{88.83} & \pmb{94.31} \\
		\cdashline{1-7}[2pt/2pt]
		\multirow{4}*{30\%} & $\times$ & $\times$ & 84.85 & 12.87 & 29.39 & 45.58\\
		~ & $\checkmark$ & $\times$ & 86.46 & 30.10 & 50.84 & 65.47 \\
		~ & $\times$ & $\checkmark$ & 86.33 & 21.66 & 39.78 & 55.97 \\
		~ & $\checkmark$ & $\checkmark$ & \pmb{86.50} & \pmb{30.30} & \pmb{51.38} & \pmb{65.84} \\
		\hline
	\end{tabular}
\end{table}

\begin{table}[!ht]
	\centering
	\caption{Comparisons between variants of DC-gra2seq on DS2, by activating two types of PEs, or not.}
	\label{tab:ablation_2pe_ds2}
	\begin{tabular}{ccccrrr}
		\hline
		\multirow{2}*{Mask} & \multirow{2}*{\makecell{Absolute\\PE}} & \multirow{2}*{\makecell{Relative\\PE}} & \multirow{2}*{$Rec$} & \multicolumn{3}{c}{$Ret$} \\ \cline{5-7}
		~ & ~ & ~ & ~ & @1 & @10 & @50 \\
		\hline
		\multirow{4}*{0\%} & $\times$ & $\times$ & 94.85 & 90.83 & \pmb{98.29} & 99.08 \\
		~ & $\checkmark$ & $\times$ & 95.36 & 91.92 & 98.04 & 99.06 \\
		~ & $\times$ & $\checkmark$ & 94.80 & 90.82 & 97.47 & 98.44 \\
		~ & $\checkmark$ & $\checkmark$ & \pmb{95.70} & \pmb{92.75} & 98.08 & \pmb{99.42} \\
		\cdashline{1-7}[2pt/2pt]
		\multirow{4}*{10\%} & $\times$ & $\times$ & 91.24 & 50.42 & 73.35 & 85.18 \\
		~ & $\checkmark$ & $\times$ & 92.45 & 66.11 & 84.01 & 90.57 \\
		~ & $\times$ & $\checkmark$ & 91.89 & 63.92 & 81.44 & 90.04 \\
		~ & $\checkmark$ & $\checkmark$ & \pmb{92.64} & \pmb{67.61} & \pmb{84.31} & \pmb{91.17} \\
		\cdashline{1-7}[2pt/2pt]
		\multirow{4}*{30\%} & $\times$ & $\times$ & 82.85 & 12.19 & 29.37 & 46.53\\
		~ & $\checkmark$ & $\times$ & 84.48 & 29.76 & 49.26 & 62.18 \\
		~ & $\times$ & $\checkmark$ & 82.95 & 29.25 & 48.76 & 61.48 \\
		~ & $\checkmark$ & $\checkmark$ & \pmb{84.62} & \pmb{29.92} & \pmb{49.26} & \pmb{62.44} \\
		\hline
	\end{tabular}
\end{table}

\begin{table}[!ht]
	\centering
	\caption{Comparisons between variants of DC-gra2seq on DS3, by activating two types of PEs, or not.}
	\label{tab:ablation_2pe_ds3}
	\begin{tabular}{ccccrrr}
		\hline
		\multirow{2}*{Mask} & \multirow{2}*{\makecell{Absolute\\PE}} & \multirow{2}*{\makecell{Relative\\PE}} & \multirow{2}*{$Rec$} & \multicolumn{3}{c}{$Ret$} \\ \cline{5-7}
		~ & ~ & ~ & ~ & @1 & @10 & @50 \\
		\hline
		\multirow{4}*{0\%} & $\times$ & $\times$ & 89.83 & 94.05 & 98.72 & 99.57 \\
		~ & $\checkmark$ & $\times$ & 89.97 & 94.94 & 98.74 & 99.71 \\
		~ & $\times$ & $\checkmark$ & 89.82 & 94.71 & 98.44 & 99.62 \\
		~ & $\checkmark$ & $\checkmark$ & \pmb{90.41} & \pmb{96.27} & \pmb{99.47} & \pmb{99.81} \\
		\cdashline{1-7}[2pt/2pt]
		\multirow{4}*{10\%} & $\times$ & $\times$ & 83.38 & 40.20 & 63.40 & 77.52\\
		~ & $\checkmark$ & $\times$ & 94.09 & 58.36 & 76.07 & 85.51 \\
		~ & $\times$ & $\checkmark$ & 83.88 & 49.40 & 69.86 & 81.77 \\
		~ & $\checkmark$ & $\checkmark$ & \pmb{85.21} & \pmb{71.66} & \pmb{86.34} & \pmb{92.35} \\
		\cdashline{1-7}[2pt/2pt]
		\multirow{4}*{30\%} & $\times$ & $\times$ & 71.07 & 5.65 & 17.40 & 32.90\\
		~ & $\checkmark$ & $\times$ & 72.52 & 18.44 & 33.39 & 47.02 \\
		~ & $\times$ & $\checkmark$ & 72.11 & 15.54 & 30.79 & 45.89 \\
		~ & $\checkmark$ & $\checkmark$ & \pmb{72.84} & \pmb{27.64} & \pmb{47.34} & \pmb{60.61} \\
		\hline
	\end{tabular}
\end{table}

\subsection{Performance gained from two types of PEs}

In this section, we verify the impacts of absolute and relative PEs introduced in DC-gra2seq. We deactivate two types of PEs by removing the matrices $\tilde{\bm P}$ or $\tilde{\bm R}$ in Eq. (\ref{eq:stack}). The experimental results for three datasets are shown in Table \ref{tab:ablation_2pe_ds1}-\ref{tab:ablation_2pe_ds3}.

With the equipments of both PEs, DC-gra2seq achieves the best performance in four variant models on controllable synthesis (Mask~=~0\%) and sketch healing (Mask~=~10\%, 30\%). Moreover, it is worth noting that the attendance of absolute PE achieves significant improvements in both $Rec$ and $Ret$. The sequential information stored in drawing orders could be beneficial for learning graphic sketch representations, and a proper usage (e.g., injecting it into graph nodes by DC-gra2seq) could be more efficient to obtain the performance gain. Besides, activating relative PE jointly with absolute PE, rather than using relative PE solo, would contribute to better performances. The cooperation between two types of PEs guides DC-gra2seq focus on not only the pairwise sequential position differences between sketch patches, but also the information about when a patch is drawn in the drawing order. Moreover, our learnable relative PE encourages DC-gra2seq to flexibly utilize the contextual distances between patches, driving to performance improvements in both tasks.

\section{Conclusions}

We have presented DC-gra2seq to encode sketch drawing orders by PE for learning context-aware graphic sketch representations. Each sketch patch is equipped with a sinusoidal absolute PE, indicating when it is drawn in a drawing order, and a learnable relative PE, encoding the contextual relationships between sketch patches. Both types of PEs are injected into graph nodes instead of edges to learn robust sketch representation against variants of sketch drawings. The attendance of PEs encourages DC-gra2seq to simultaneously consider the visual patterns from the canvas, the sequential positions from drawing orders and the unseen contextual information between patches, which benefits sketch representation learning verified on controllable sketch synthesis and sketch healing.

\section{Acknowledgments}

This work was supported by the Fundamental Research Funds for the Central Universities (2232024D-28) and the National Natural Science Foundation of China (62406064).

\bibliographystyle{elsarticle-num} 
\bibliography{reference}

\end{document}